\documentclass[letterpaper, 10 pt, conference]{ieeeconf}  

\let\labelindent\relax

\IEEEoverridecommandlockouts                              

\usepackage[utf8]{inputenc}
\usepackage{lipsum}
\usepackage{multicol}
\usepackage{graphicx}
\usepackage{enumitem}
\usepackage{amsmath}
\usepackage{multirow}
\usepackage{gensymb}
\usepackage{fancyhdr}

\title{\LARGE \bf
Autonomous Tissue Scanning Under Free-Form Motion for Intraoperative Tissue Characterisation}

\author{Jian Zhan$^{*,1}$, Jo\~ao Cartucho$^{*,1}$ and Stamatia Giannarou$^{1}$
\thanks{$^{*}$These authors contributed equally to the work.}
\thanks{Jian Zhan:
        {\tt\small j.zhan18@imperial.ac.uk}}%
        \thanks{Jo\~ao Cartucho:
        {\tt\small j.cartucho19@imperial.ac.uk}}%
\thanks{$^{1}$The Hamlyn Centre for Robotic Surgery, Imperial College London, London SW7 2AZ, UK}%
}

\graphicspath{{figures/}}

\begin{document}

\maketitle
\thispagestyle{empty}
\pagestyle{empty}


\begin{abstract}


In Minimally Invasive Surgery (MIS), tissue scanning with imaging probes is required for subsurface visualisation to characterise the state of the tissue. However, scanning of large tissue surfaces in the presence of deformation is a challenging task for the surgeon. Recently, robot-assisted local tissue scanning has been investigated for motion stabilisation of imaging probes to facilitate the capturing of good quality images and reduce the surgeon's cognitive load. Nonetheless, these approaches require the tissue surface to be static or deform with periodic motion. To eliminate these assumptions, we propose a visual servoing framework for autonomous tissue scanning, able to deal with free-form tissue deformation. The 3D structure of the surgical scene is recovered and a feature-based method is proposed to estimate the motion of the tissue in real-time. A desired scanning trajectory is manually defined on a reference frame and continuously updated using projective geometry to follow the tissue motion and control the movement of the robotic arm. The advantage of the proposed method is that it does not require the learning of the tissue motion prior to scanning and can deal with free-form deformation. We deployed this framework on the da Vinci\textsuperscript{\textregistered} surgical robot using the da Vinci Research Kit (dVRK) for Ultrasound tissue scanning. Since the framework does not rely on information from the Ultrasound data, it can be easily extended to other probe-based imaging modalities. 
\end{abstract}


\section{INTRODUCTION}

Minimally Invasive Surgery (MIS) has rapidly gained popularity in the last decades and is now the gold standard for procedures, such as urology and congenital heart disease surgery.
To overcome the fundamental challenges of MIS in terms of constraint workspace and poor triangulation, while robotic technology has been applied to improve surgical dexterity and comfort, laparoscopic systems have been utilised to provide visualisation of the surgical environment.

Recently, autonomy has been introduced in robotic surgery, allowing robotic surgical systems to complete specific tasks without supervision~\cite{moustris2011evolution}. This automation enables precise execution of surgical tasks even under challenging conditions such as soft tissue deformation with reduced surgical workload and increased patient safety ~\cite{hennersperger2016towards, shademan2016supervised}. Surgical automation is still at its infancy and so far has focused on the execution of low-level surgical tasks such as cutting and suturing divided into sub-tasks, including needle grasping and selection of incision port. Cochlear implants have also greatly benefited from autonomy, considering the limited operational space and required precision.

Recent advances in intraoperative imaging such as intraoperative pick-up Ultrasound and probe-based Confocal Laser Endomicroscopy (pCLE) have enabled in vivo, in situ tissue characterisation. Despite the level of detail provided by these modalities, the use of miniaturised imaging probes makes a systematic examination of large and deformable tissue surfaces a challenging task in practice, due to its requirements of a high level of precision and stability. Automating tissue scanning with such imaging probes allows not only to improve the quality of the captured data, but also to reduce the operational time and decrease surgical workload. Current approaches to robot-assisted local tissue scanning rely on the assumption that the tissue is static or moving with periodic motion.

This work aims to propose a visual servoing framework for autonomous tissue scanning, optimising probe-tissue contact and compensating for free-form tissue motion. The proposed framework advances state-of-the-art autonomous tissue scanning methods by eliminating the requirement for learning the tissue motion before scanning and dealing with free-form tissue deformation in real-time. The proposed framework has been validated on phantom data.
This paper is structured as follows: Section II reviews the state-of-the-art, Section III describes the methodology, Section IV, the experimental setup and analyzes the results, and Section V presents our conclusions.


\section{RELATED WORK}

Current approaches to robot-assisted tissue scanning with imaging probes have focused on applications using Ultrasound and pCLE as imaging modalities. The above modalities should firmly touch the tissue surface while closely following the tissue motion to capture good quality imaging data.


Previous studies on autonomous Ultrasounds scanning have focused on image-guided needle insertion~\cite{zettinig20173d}, Focused Assessment with Sonography for Trauma (FAST)~\cite{virga2016automatic}, tumour detection~\cite{zhang2017motion} and vessel tracking~\cite{merouche2015robotic}. In the above visual servoing frameworks information either from the Ultrasounds data \cite{Inrianadeau2016moments} or the endoscopic camera \cite{pratt2015autonomousultrasound-guidedtissuedissection} has been used as visual feedback for probe positioning or sweeping scanning. Recently, force feedback has also been included for Ultrasounds-guided flexible needle insertion with haptic feedback \cite{Inriachevrie2019real}. For applications where the robot trajectory needs to be planned in advance, the 3D structure has been recovered using stereo cameras~\cite{zhang2017motion}, RGB-D cameras~\cite{huang2018robotic} or combination of RGB-D and preoperative MRI images~\cite{hennersperger2016towards}. 







Previous studies on autonomous scanning with pCLE have addressed the problem of tissue contact by using force feedback from force sensors~\cite{zhang2017autonomous} and stereo vision for pose estimation. Optical Coherence Tomography (OCT) has also been employed~\cite{zhang2017macro} to measure the distance between the probe and the tissue and fuse OCT and pCLE data for enhanced visualisation.~\cite{varghese2017framework} proposed a deep learning framework with image criteria of blur as input, to classify the position of the probe as too close, too far or at the right distance from the tissue. The approach proposed by ~\cite{triantafyllou2018framework} also controlled the probe/tissue distance based only on information from the pCLE data, by applying the Crete-Roffet Blur Metric (CRBM) along with fuzzy logic. Most of the studies have adopted the mosaicking algorithm discussed in~\cite{rosa2012building}, which has been updated by the method proposed in~\cite{giataganas2018intraoperative}.




It is essential to track and estimate the motion of the tissue in real-time to develop an intraoperative visual servoing framework robust to tissue deformation. Recent work has focused on the tracking of a target lesion to improve Ultrasounds-guided breast biopsy~\cite{nikolaev2019ultrasound}. Similarly, in~\cite{op2011predicting} the displacement of a subsurface target is predicted due to interactions between the surgical tool and the soft-tissue.

Previous works on tissue tracking~\cite{daly2010fusion, de2010augmented, teber2009augmented} added fiducial markers or synthetic features to the exposed tissue surface, in order to monitor the dynamic 3D displacement of the tissue in real-time. To deal with occlusions that occur during long-term tracking the assumption of a static scene has been used~\cite{grasa2011ekf} or robotic occlusion avoidance has been proposed~\cite{wang2015robot}. Tracking the tissue becomes more challenging when we take into consideration that the tissue may have a homogeneous texture, and consequently, only a few distinct features are available for tracking the tissue's motion. Previous work has shown that in MIS we can track soft tissue using salient features tailored for MIS applications~\cite{giannarou2016vision},~\cite{giannarou2012probabilistic} or increase the number of features being matched correctly, between different images, by fusing different feature detectors~\cite{mountney2007probabilistic}. However, the fusion of descriptors is more computationally expensive, which restricts its direct use in real-time applications such as tissue scanning. 





\section{METHODOLOGY}

\subsection{Framework Overview}


The framework is established upon the da Vinci Research Kit (dVRK)\cite{kazanzides-chen-etal-icra-2014dvrk}. Components include the stereo endoscopic camera, a Patient Side Manipulator (PSM) linked to a dVRK controller, and Cadiere forceps (Intuitive Surgical Inc., Sunnyvale, CA, USA). UTS-533 linear array Ultrasounds probe is used, connecting to a ProSound Alpha 10 (Hitachi Aloka Medical Ltd., Tokyo, Japan). A KeyDot\textsuperscript{\textregistered} marker (Key Surgical Inc., Eden Prairie, MN, USA) with 7 x 3 asymmetric circular pattern is adhered to the Ultrasounds probe. The software system is built with the Robot Operating System (ROS), facilitating parallel computation and the communication with the stereo endoscopic camera and dVRK controllers. 

Our focus was to minimise the required setup time and avoid to introduce additional hardware (e.g., RGB-D cameras) to our dVRK system, to allow seamless integration of our proposed autonomous tissue scanning in the Operating Theatre. Therefore, our visual feedback is based on RGB images from the endoscopic stereo camera. 

\subsection{Visual Servoing}
\label{subsec:Visual_servoing}
A Position Based Visual Servoing (PBVS) scheme is applied in the proposed framework, which can be expressed as:
\begin{equation}
	e(t)=s(m(t),a)-s* 
\label{eq:vs_general}
\end{equation}
where, $s$ and $s*$ are the current state and the desired state respectively, $e$ is the error to be minimised, $m(t)$ is the visual measurement and $a$ represents extra knowledge \cite{chaumette2006visualTut1}. In this framework, the visual state is the marker pose with respect to the camera, and by representing them in homogeneous matrix form, (\ref{eq:vs_general}) can be represented as: 
\begin{equation}
    ^{M}T_{M*} =\ (^{C}T_{M})^{-1} \ ^{C}T_{M*}.
\label{eq:MTM*}
\end{equation}
where, the notation $^{M}T_{M*}$ means the homogeneous transform from the desired marker pose (M*) to the current marker pose (M), where $*$ indicates the desired state. Similarly, $^{C}T_{M*}$ is the transform from the desired marker pose to the camera and $ (^{C}T_{M})^{-1}= \, ^{M}T_{C}$ is the transform from the camera to the current marker pose, which is obtained by employing the marker tracking method proposed in \cite{pratt2015robustUS} . All the transformations used in this framework are illustrated and described in Fig. \ref{fig:transforms}. The dVRK controllers allow controlling the PSM arm directly in Cartesian space by sending the command with the transforms from its end-effector to the base, namely $^{B}T_{E*}$, which can be calculated with the following equation:
\begin{figure}[tp]
    \centering
    \includegraphics[width=7cm]{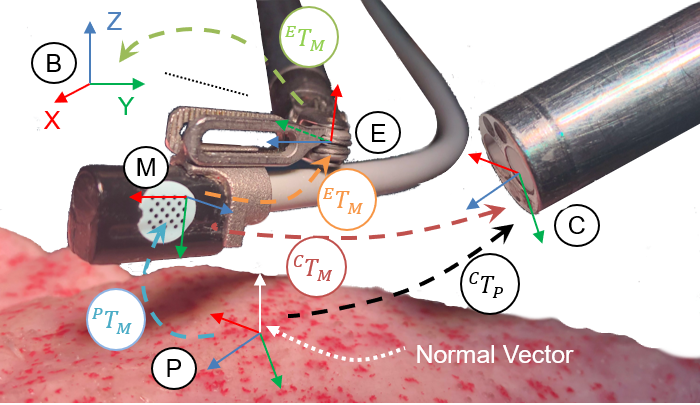}
    \caption{The coordinate systems and their transformations. Legend: Robot B - Base,  C - Endoscopic Camera, E - PSM End effector, M - Marker, and P - Tissue/Probe Contact Point.}
    \label{fig:transforms}
\end{figure}
\begin{figure}[pth]
    \centering
    \includegraphics[width=8cm]{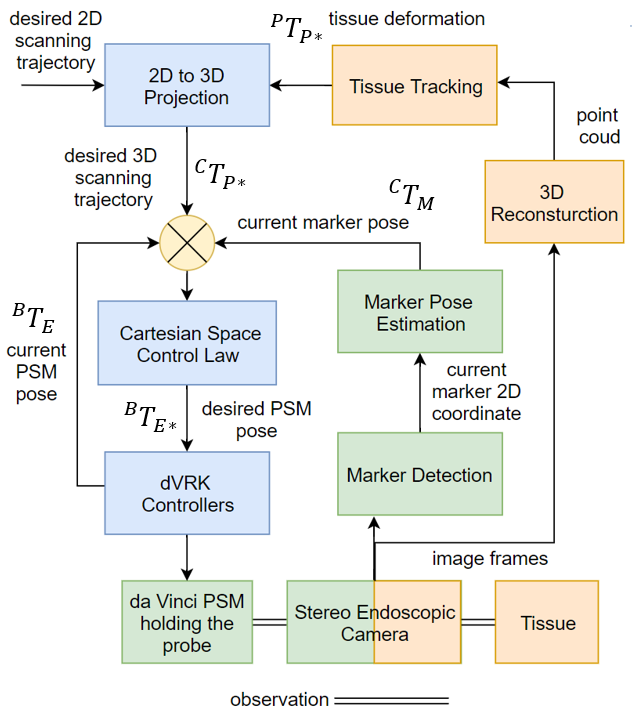}
    \caption{The multi-loop control architecture, where the inner loop is visual feedback from the marker (in green) (25HZ), and outer loop (in orange) yields the feedback of tissue motion (10HZ).}
    \label{fig:contorl_loop}
\end{figure}
\begin{equation}
    ^{B}T_{E*} =\ ^{B}T_{E} \ ^{E}T_{E*}
    \label{BTE*}
\end{equation}
where, $^{B}T_{E}$ is the current end-effector pose to the base, which is measured by the PSM's encoders. $^{E}T_{E*}$ can be expanded with (\ref{eq:MTM*}) through the transform chain forming a control law as:
\begin{equation}
    ^{B}T_{E*} =\ ^{B}T_{E} \ ^{E}T_{M} \ ^{M}T_{C} \ ^{C}T_{P} \ ^{P}T_{P*} \ ^{P*}T_{M*} \ ^{M*}T_{E*}
\label{eq:contorllaw}
\end{equation}
where, $^{E}T_{M} = (^{M*}T_{E*})^{-1}$ is a rigid transform between the marker and the PSM end-effector which is obtained with hand-eye calibration \cite{shah2013solving}. $ ^{C}T_{P}$ is transform from the tissue point to the camera , attained through 3D reconstruction. $^{P*}T_{M*}$ describe the transformation from the desired marker pose to the desired contact point when the probe makes contact with the point. The orientation is constrained by the normal vector and the orientation of the marker that always faces to the camera. $^{P}T_{P*}$ represents the transform from the initial tissue contact point to the updated point through tissue tracking. Our proposed multi-loop control architecture is shown in Fig. \ref{fig:contorl_loop}.

\begin{figure*}[!ht]
  \centering
  \includegraphics[width=\textwidth]{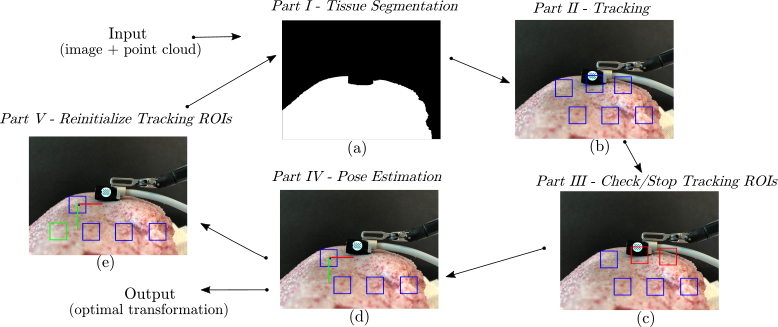}
  \caption{The tissue tracking algorithm is a five-part loop that outputs an optimal transformation (translation + rotation) using as input an image + point cloud. The algorithm starts by conducting a binary segmentation of the input image (a) to label the tissue's pixels, as shown in white. Then, it tracks a sparse set of ROIs, as shown in blue (b). Then, it checks for occlusions in the ROIs. If an occlusion is detected, for example an occlusion of the robotic arm, it ``Stops'' tracking that ROI, as shown in red (c). Using the ROIs in blue (d) the algorithm estimates the current pose of the tissue by calculating optimal initial-to-current transformation. Finally, using the optimal transformation, it tries to ``Reinitialize'' the ROIs previously ``Stopped'' on a previous loop, shown in green (e).}
  \label{fig:pose_estimation_loop}
\end{figure*}
\subsection{3D Reconstruction} 
\label{subsec:3d_reconstruction}
In our proposed robotic tissue scanning framework, the imaging probe (e.g., an Ultrasounds probe) is moved over the tissue surface following a predefined 3D scanning trajectory. To deal with deformation, this trajectory is adjusted to the tissue surface to ensure that the probe is at the right distance and perpendicular to the tissue surface. Therefore, the recovery of the 3D structure of the tissue is a key module of this framework.

The 3D reconstruction module computes a point cloud and the normal vectors of the surface using as input the rectified and undistorted stereo images. The Efficient Large-scale Stereo Matching (ELAS) \cite{geiger2010efficient} method is adopted to estimate the disparity, whose effectiveness and efficiency in the medical context has been verified in \cite{zhang2017autonomous} and \cite{zhang2017macro}. The left camera is defined as the reference coordinate frame and therefore, the disparity corresponding to the left camera is applied to generate the point cloud. After obtaining the point cloud, the surface normal of each point is estimated by computing the first derivative at two orthogonal directions and taking the cross product of them. 


\subsection{Tissue Tracking}
\label{subsec:tissue_tracking}


Our proposed framework requires the robot to move a probe along a 3D scanning trajectory in the presence of tissue deformation. Therefore the scanning trajectory needs to be continuously corrected to adapt to the tissue shape in real-time.
To that end, we calculate the optimal initial-to-current rigid transformation of the tissue, where by initial we refer to the tissue reference point cloud (when the scanning starts) and by current we refer to the most recent point cloud (the one computed in real-time).
The estimated rigid transformation is applied to every point of the reference trajectory to update the scanning trajectory. The rigid transformation assumption is acceptable since we are scanning a relatively small tissue area which preserves its shape and size during scanning.


To estimate the above rigid transformation, the tissue surface needs to be tracked along time to establish point correspondences. In our work, a sparse set of Regions Of Interest (ROIs) distributed over the tissue surface are tracked and their motion, between consecutive frames, is estimated. In this module, we used the ORB feature detector~\cite{rublee2011orb} and the MedianFlow tracking~\cite{kalal2010forward} due to their real-time performance and effectiveness.

The input for our tracking algorithm is a rectified image captured by the robot's stereo camera and the point cloud corresponding to that image which is calculated by our 3D reconstruction module (Sec:~\ref{subsec:3d_reconstruction}). Before starting the tissue tracking, the algorithm initializes a sparse set of ROIs over the tissue surface. Those ROIs are rectangles of size $20 \times 25$ pixels and are positioned in the areas of the tissue with the more distinctive features.


The tissue pose estimation results from the following five-part looping algorithm, which is illustrated in Fig. \ref{fig:pose_estimation_loop}:

\subsubsection{Part I - Tissue Segmentation}

In Part I, a binary image segmentation method is applied to distinguish the pixels that belong to the tissue from the rest of the scene. Pixels that do not belong to the tissue (e.g., pixels of the robotic arm) are not relevant for estimating the pose of the tissue.
 In specific, the algorithm does a colour-based segmentation using the HSV colour space.

\subsubsection{Part II - Tracking}


In Part II, the motion of each ROI is estimated so that when the tissue moves, the ROIs also move accordingly. By tracking a sparse set of ROIs in consecutive 2D input images, the algorithm implicitly tracks a set of 3D points which are extracted from the point cloud at the centre of each ROI. These 3D points are used in Part IV for estimating the current pose of tissue.



\subsubsection{Part III - Check/Stop Tracking ROIs}


After tracking (Part II), the algorithm checks for occlusions. An occlusion is detected when the percentage of tissue pixels inside an ROI is smaller than a given threshold which is empirically determined. Each pixel is labelled as tissue or not using the segmented image from Part I. If an occlusion is detected or if the ROI's visual appearance (intensity histogram + ORB features) changes significantly, then the algorithm ``Stops'' tracking that ROI, and it will not be used for estimating the tissue's pose.


\subsubsection{Part IV - Pose Estimation}

The algorithm then estimates the tissue pose by calculating the optimal initial-to-current rigid transformation.


The algorithm uses the RANSAC iterative process to take into consideration the presence of noise and outliers. It randomly samples a set of three ROIs to estimate the optimal initial-to-current transformation. Given all the possible optimal transformations (one for each combination of three ROIs), we choose the one that maximises the numbers of inliers and minimises the distance between the predicted and the current positions of the 3D points (the points being tracked at the centre of each ROI).

To calculate the optimal transformation, given each combination of three ROI, we find the least-squares fitting of two sets of corresponding 3D points using a Singular Value Decomposition (SVD). First, we calculate the centroids of the initial and current point sets:

\begin{equation}
p_{initial} = 
\begin{bmatrix}
  X_{initial} \\
  Y_{initial} \\
  Z_{initial} \\
\end{bmatrix}
   \quad\text{and}\quad 
p_{current} =
 \begin{bmatrix}
  X_{current} \\
  Y_{current} \\
  Z_{current} \\
\end{bmatrix}
\end{equation}

\begin{equation}
c_{initial} = \frac{1}{3}\sum_{k=1}^{3} p_{initial}^{k}
 \quad\text{and}\quad 
 c_{current} = \frac{1}{3}\sum_{k=1}^{3} p_{current}^{k}
\end{equation}

where, $p_{initial}$ and $p_{current}$ are the initial and current points being tracked, and $c_{initial}$ and $c_{current}$ are the centroids of those point sets.

Given the centroids we can then find the optimal rotation by finding the SVD of $H$, a $3\times3$ matrix calculated by:

\begin{equation}
H = \sum_{k=1}^{3} (p_{initial}^{k} - c_{initial}) (p_{current}^{k} - c_{current})
\end{equation}

\begin{equation}
\begin{bmatrix}
  U & S & V
\end{bmatrix}
= SVD(H)
 \quad\text{and}\quad 
r = V U^{T}
\end{equation}

where, $r$ is the optimal rotation. If the determinant of $r$ is negative, we then multiply the third column of $V$ by $-1$ and then recompute $r$.

Then, we calculate the translation by:

\begin{equation}
t = c_{current} - r \times c_{initial}
\end{equation}

where, $t$ is the optimal translation.



\subsubsection{Part V - Reinitialize Tracking ROIs}

Using the optimal transformation from Part IV, the algorithm estimates where the ROIs that were ``Stopped'' being tracked in the previous loops would be in the current image by projecting the initial 3D point to the current 2D image:

\begin{equation}
 s \times
\begin{bmatrix}
 u_{current} \\
 v_{current} \\
 1 \\
\end{bmatrix}
= K_{left} \times M \times p_{initial}
\end{equation}

where, $M$ is the $4 \times 4$ optimal transformation calculated in Part IV,  $K_{left}$ is the $3 \times 4$ projection matrix in the rectified coordinate system for the left camera of the stereo camera and, $u_{current}$ and $v_{current}$ are the pixel coordinates of the estimated ROI's position in the image.

Then, we compare the visual appearance (intensity histogram + ORB features) of the estimated ROI with its reference and if we have a match then we ``Reinitialize'' that ROI.

Finally, the algorithm goes back to Part I with the new image and corresponding point cloud, and it repeats the loop until the scanning has been completed.

With the optimal transformation, the robot can scan a tissue that is unpredictably moving in different directions, which is the main novelty of this paper.
\section{EXPERIMENTS AND RESULTS}
\subsection{Experimental Setup}


Our framework was implemented in C++ on a PC with an Intel Core (i7-3770) and 16 GB RAM. In the current implementation, the framework runs on images with a 576 $\times$ 720 resolution. The visual feedback from the marker and the control of the PSM run at 25 frames per second and our ``3D reconstruction $+$ tissue tracking'' loop at approximately 10 frames per second.

To evaluate the proposed framework, we used a silicon liver phantom which was deformed using a motorised platform controlled by three Strada\textsuperscript{\textregistered} motors. Each motor controls the motion along one of the three orthogonal directions. The platform is programmable and therefore, similarly to~\cite{zhang2017motion}, we selected three periodic respiratory motions (Profile 1, Profile 2 and Profile 3) corresponding to different breathing speeds and amplitudes as shown in Table \ref{tab:motionProfile}. Furthermore, we introduced random free-form deformations to those motions to test if the tissue is successfully tracked when it arbitrarily moves in different directions.

\begin{table}[htb]
\begin{center}
\caption{Respiratory Deformations for the Moving Platform}
\label{tab:motionProfile}
\begin{tabular}{|r|c|c|c|}
    \hline
    Deformation & \textbf{Profile 1} & \textbf{Profile 2} & \textbf{Profile 3} \\
    \hline
    Period & 3 s & 5 s & 5 s \\
    Amplitude & 3 mm & 3 mm & 5 mm \\ 
    \hline
\end{tabular}
\label{motionprofile}
\end{center}
\end{table}


\subsection{Accuracy of Visual Servoing}



This experiment is designed to evaluate how well the probe is physically following the desired pose in real-time. In this experiment, our ground-truth is the pose of the marker. To validate this experiment we calculated the mean and standard deviation of the error between the current and the desired marker's pose. 


For this experiment, we conducted a total of ten trials, three for each motion Profile (Tab: \ref{tab:motionProfile}) and one for free-form motion. Sinusoidal waveform has been generated so that trajectories were created based on motion Profile 1, 2 and 3 from Table \ref{motionprofile} (e.g. $1.5sin(\frac{2\pi}{3}t)$ for the Profile 1). The results are tabulated in Table \ref{visualservoingerror}. It shows that, with visual  servoing the robot was able  to  achieve  on  average   0.75  mm  for  the transnational error and 0.62\degree  for the rotational error.

\begin{table}
\begin{center}
\caption{Visual Servoing Error (Mean $\pm$ STD)} \label{tab:visual_srvoing_error} 
\begin{tabular}{ |c|c|c|c|c| } 
\hline
\multicolumn{2}{|c|}{Deformation}  & Translation Error (mm) & Rotation Error (\degree)\\
\hline
\multicolumn{2}{|c|}{\textbf{Free-form}} & 1.796 $\pm$ 0.728 & 1.018 $\pm$ 0.604 \\
\hline
\multirow{3}{4em}{\textbf{Profile 1}} & x & 0.683 $\pm$ 0.222 & 0.505 $\pm$ 0.212\\ 
& y & 0.696 $\pm$ 0.348 & 0.578 $\pm$ 0.176\\ 
& z & 0.662 $\pm$ 0.308 & 0.376 $\pm$ 0.171\\ 
\hline
\multirow{3}{4em}{\textbf{Profile 2}} & x & 0.567 $\pm$ 0.166 & 0.513 $\pm$ 0.217\\ 
& y & 0.544 $\pm$ 0.248 & 0.552 $\pm$ 0.195\\ 
& z & 0.519 $\pm$ 0.222 & 0.383 $\pm$ 0.192\\ 
\hline
\multirow{3}{4em}{\textbf{Profile 3}} & x & 0.752 $\pm$ 0.219 & 0.874 $\pm$ 0.377\\ 
& y & 0.624 $\pm$ 0.290 & 0.828 $\pm$ 0.325\\ 
& z & 0.699 $\pm$ 0.267 & 0.570 $\pm$ 0.210\\
\hline
\multicolumn{2}{|c|}{\textbf{Total}} & \textbf{0.754 $\pm$ 0.302} & \textbf{0.620 $\pm$ 0.268}\\
\hline
\end{tabular}
\label{visualservoingerror}
\end{center}
\end{table}

\subsection{Accuracy of Tissue Tracking}

To quantify the accuracy and precision of the tissue tracking, we calculated the mean and standard deviation of the error between the estimated and the ground truth tissue poses at a given time instance. The ground truth pose was measured by using an ArUco fiducial marker~\cite{garrido2014automatic}, placed on the tissue surface. The ArUco marker did not influence the result of the estimated pose since, during validation, as we ensured that no ROI of the tracking algorithm contained pixels of the ArUco marker. ArUco was only applied in this experiment for obtaining ground truth and was removed in other experiments. 

Firstly, we conducted a trial where a research participant held the tissue phantom in his hand and rotated it slightly around the three-axis so that we could compare the rotations measured by the ArUco with the ones predicted by the tracking algorithm. The mean error was of 2\degree.

Secondly, we conducted twelve trials, four for each motion Profile (Tab: \ref{tab:motionProfile}). Using the moving platform, we moved the tissue along the three-axis (x, y and z), whose results are tabulated in Table \ref{tab:tissue_tracking_error}. In that table, the ``free-form'' refers to a random combination of movements along the x, y and z-axis. In all the trials the tissue was moving for a total of 30 seconds.

We can see from the table that in all the trials, the mean translation error was smaller than 1 mm, and the mean rotational error was smaller than 3\degree.
Overall, the mean translation error was of 0.7 mm, and the mean rotational error was 1.4\degree.







\begin{table}[t]
\begin{center}
\caption{Tissue Tracking Error (Mean $\pm$ STD)} \label{tab:tissue_tracking_error}
\begin{tabular}{ |c|c|c|c|c| } 
\hline
\multicolumn{2}{|c|}{Deformation}  & Translation Error (mm) & Rotation Error (\degree)\\
\hline
\multicolumn{2}{|c|}{\textbf{In-hand rotation}}& - & 2.120 $\pm$ 1.880 \\
\hline
\multirow{4}{4em}{\textbf{Profile 1}} & x & 0.691 $\pm$ 0.437 & 1.172 $\pm$ 0.589\\ 
& y & 0.678 $\pm$ 0.437 & 0.989 $\pm$ 0.464\\ 
& z &  0.580 $\pm$ 0.261 & 0.915 $\pm$ 0.452\\ 
& free-form & 0.521 $\pm$ 0.214 & 1.504 $\pm$ 0.769\\ 
\hline
\multirow{4}{4em}{\textbf{Profile 2}} & x & 0.854 $\pm$ 0.375 & 1.183 $\pm$ 0.506\\ 
& y & 0.966 $\pm$ 0.392 & 1.597 $\pm$ 0.669\\ 
& z & 0.878 $\pm$ 0.308 & 2.160 $\pm$ 0.845\\ 
& free-form & 0.647 $\pm$ 0.336 & 0.997 $\pm$ 0.497\\ 
\hline
\multirow{4}{4em}{\textbf{Profile 3}} & x & 0.706 $\pm$ 0.352 & 1.134 $\pm$ 0.645\\ 
& y & 0.573 $\pm$ 0.255 & 1.229 $\pm$ 0.613\\ 
& z & 0.696 $\pm$ 0.291 & 1.600 $\pm$ 0.694\\ 
& free-form & 0.739 $\pm$ 0.312 & 1.376 $\pm$ 0.615\\ 
\hline
\multicolumn{2}{|c|}{\textbf{Total}} & \textbf{0.711 $\pm$ 0.315} & \textbf{1.383 $\pm$ 0.711}\\
\hline
\end{tabular}
\end{center}
\end{table}






\subsection{Ultrasound Stability}

Our framework's ultimate goal is to be able to scan a region of the tissue stably. Therefore, the comparison between different Ultrasound frames is a way of evaluating our framework. In specific, both a qualitative and quantitative evaluation of Ultrasound images were conducted. In the quantitative evaluation, we continuously scan the same position on the tissue surface while the tissue is deforming, and compare the initial Ultrasound image with the consecutive ones in real-time by calculating the Normalised Cross-Correlation (NCC). A higher NCC score indicates a higher similarity between the Ultrasound images.

\begin{figure}[tbh]
    \centering
    \includegraphics[width=8cm]{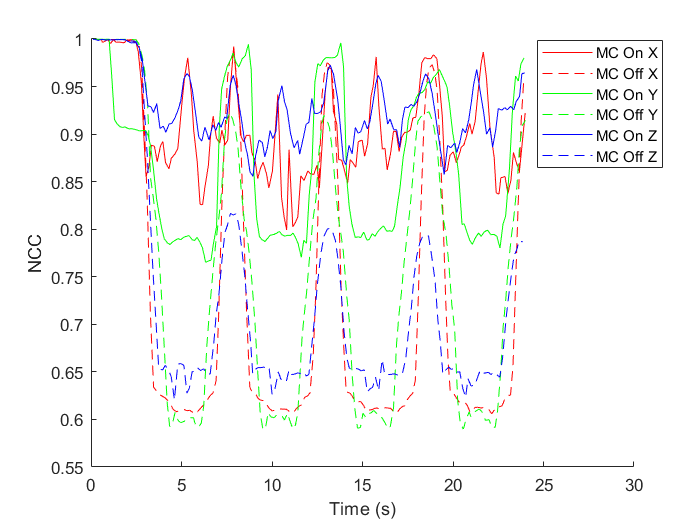}
    \caption{In this graph, the NCC score is calculated by comparing the initial Ultrasound image with the consecutive ones, with and without Motion Compensation (MC), On/Off. In specific, the tissue is deforming with the characteristics of the motion Profile 3 in the x, y and z-direction. From the graph, we can see that independently of the direction, when the motion compensation is On the NCC score is significantly higher, indicating that the probe is scanning the same region of the tissue.}
    \label{NCC}
\end{figure}

\begin{figure}[tbh]
    \centering
    \includegraphics[width=8cm]{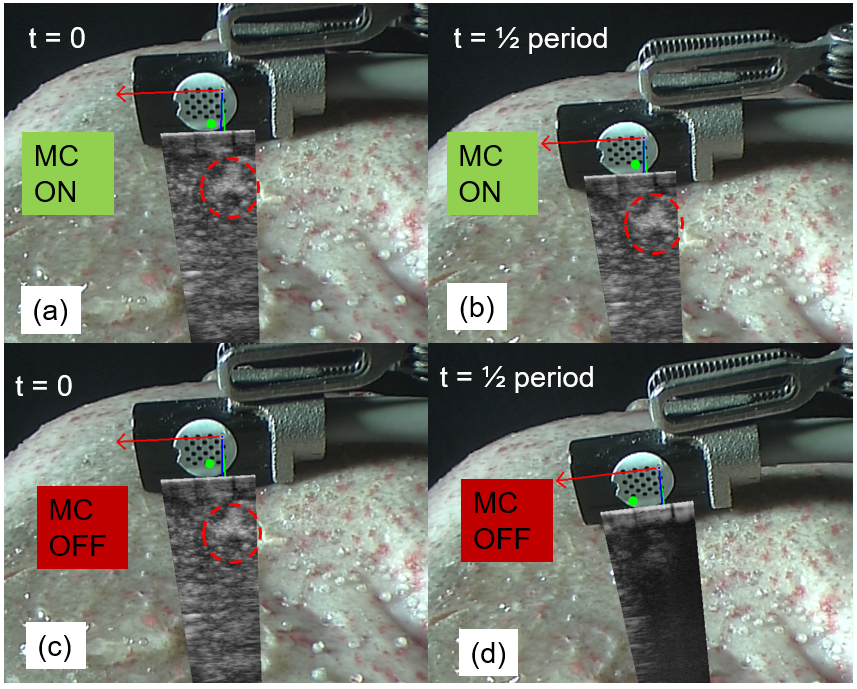}
    \caption{The Ultrasound is overlaid with the stereoscopic image in real-time to create an Ultrasound-augmented image. This augmentation allows us to validate our framework qualitatively. In both figures (a) and (c) the probe starts at the same initial pose. From (a) to (c), we can see that the Ultrasound images are very similar, and therefore, the robot is scanning the target successfully due to our Motion Compensation (MC). On the contrary, from (c) to (d), we can see that without the MC, the robot is not able to successfully scan the target position. Note that an Ultrasound image turns fully black when the probe-tissue contact is lost.}
    \label{MCOnOff}
\end{figure}

The results in Fig. \ref{NCC} show that the NCC between the compared Ultrasound images is significantly higher when motion compensation is included which verifies that our framework is able to successfully scan a tissue surface under free-form deformation. An example of qualitative result is demonstrated in Fig. \ref{MCOnOff}.





\subsection{Discussion}

The proposed framework demonstrates its ability to compensate for period and free-form tissue deformation while maintaining stable Ultrasound scanning. According to the experimental results, our framework tracks a tissue with a translation mean error of 0.7mm and  rotation mean error of 1.3\degree. We believe that this error is mostly due to noise in the estimated 3D point cloud. The experiment on visual servoing accuracy gives an error of 0.96mm and 0.77\degree. One cause of it is that the delay between the PSM receiving commands and executing the action. Every time the dVRK receives a pose command in Cartesian space, it needs to solve the inverse kinematics to determine its joint motion, which delays the action. Directly control the PSM by its joint may improve the delay, and nevertheless, the scanning trajectory is defined in Cartesian space. How controlling the robot in joint space will affect the scanning accuracy requires further research.

The computational power also limits the performance of the framework, as it is currently running on a CPU only. A comprehensive solution for this problem is to accelerate the computationally expensive modules (3D reconstruction and Tissue Tracking) with GPU computations.

For future work, we plan on validating our framework with in vivo data and perform a more thorough performance evaluation of the visual servoing component. Additionally, we plan on adding a module for reconstructing a 3D model of a tumour's surface using as input the Ultrasound images. The addition of a tumour reconstruction module is expected to benefit this work significantly since it will allow us to evaluate the entire framework by comparing the predicted volume of a tumour with a ground-truth volume obtained from a CT scan.

\section{CONCLUSIONS}


In this paper, we proposed a framework for accurate and efficient tissue scanning. Our tissue scanning framework is fully automatic, and it does not require any preoperative data about the shape of the tissue to be scanned. The main advantage of our framework is that it can deal with free-form tissue deformations which eliminate the periodic motion assumption used by other state-of-the-art methods. Our framework does not learn the motion before scanning; instead, it estimates the tissue's pose by calculating the optimal initial-to-current transformation of the tissue. This framework was deployed on the da Vinci\textsuperscript{\textregistered} robot and validated using a phantom of the liver.
\section{ACKNOWLEDGEMENTS}

The authors are grateful for the support from the NIHR Imperial BRC (Biomedical Research Centre) and the Royal Society (UF140290).


\addtolength{\textheight}{-0cm}   


\bibliographystyle{IEEEtran}
\bibliography{IEEEabrv,files/references} 

\begin{thebibliography}{10}
\providecommand{\url}[1]{#1}
\csname url@rmstyle\endcsname
\providecommand{\newblock}{\relax}
\providecommand{\bibinfo}[2]{#2}
\providecommand\BIBentrySTDinterwordspacing{\spaceskip=0pt\relax}
\providecommand\BIBentryALTinterwordstretchfactor{4}
\providecommand\BIBentryALTinterwordspacing{\spaceskip=\fontdimen2\font plus
\BIBentryALTinterwordstretchfactor\fontdimen3\font minus
  \fontdimen4\font\relax}
\providecommand\BIBforeignlanguage[2]{{%
\expandafter\ifx\csname l@#1\endcsname\relax
\typeout{** WARNING: IEEEtran.bst: No hyphenation pattern has been}%
\typeout{** loaded for the language `#1'. Using the pattern for}%
\typeout{** the default language instead.}%
\else
\language=\csname l@#1\endcsname
\fi
#2}}

\bibitem{moustris2011evolution}
G.~P. Moustris, S.~C. Hiridis, K.~M. Deliparaschos, and K.~M. Konstantinidis,
  ``Evolution of autonomous and semi-autonomous robotic surgical systems: a
  review of the literature,'' \emph{The international journal of medical
  robotics and computer assisted surgery}, vol.~7, no.~4, pp. 375--392, 2011.

\bibitem{hennersperger2016towards}
C.~Hennersperger, B.~Fuerst, S.~Virga, O.~Zettinig, B.~Frisch, T.~Neff, and
  N.~Navab, ``Towards mri-based autonomous robotic us acquisitions: a first
  feasibility study,'' \emph{IEEE transactions on medical imaging}, vol.~36,
  no.~2, pp. 538--548, 2016.

\bibitem{shademan2016supervised}
A.~Shademan, R.~S. Decker, J.~D. Opfermann, S.~Leonard, A.~Krieger, and P.~C.
  Kim, ``Supervised autonomous robotic soft tissue surgery,'' \emph{Science
  translational medicine}, vol.~8, no. 337, pp. 337ra64--337ra64, 2016.

\bibitem{zettinig20173d}
O.~Zettinig, B.~Frisch, S.~Virga, M.~Esposito, A.~Rienm{\"u}ller, B.~Meyer,
  C.~Hennersperger, Y.-M. Ryang, and N.~Navab, ``3d ultrasound
  registration-based visual servoing for neurosurgical navigation,''
  \emph{International journal of computer assisted radiology and surgery},
  vol.~12, no.~9, pp. 1607--1619, 2017.

\bibitem{virga2016automatic}
S.~Virga, O.~Zettinig, M.~Esposito, K.~Pfister, B.~Frisch, T.~Neff, N.~Navab,
  and C.~Hennersperger, ``Automatic force-compliant robotic ultrasound
  screening of abdominal aortic aneurysms,'' in \emph{2016 IEEE/RSJ
  International Conference on Intelligent Robots and Systems (IROS)}.\hskip 1em
  plus 0.5em minus 0.4em\relax IEEE, 2016, pp. 508--513.

\bibitem{zhang2017motion}
L.~Zhang, M.~Ye, S.~Giannarou, P.~Pratt, and G.-Z. Yang, ``Motion-compensated
  autonomous scanning for tumour localisation using intraoperative
  ultrasound,'' in \emph{International Conference on Medical Image Computing
  and Computer-Assisted Intervention}.\hskip 1em plus 0.5em minus 0.4em\relax
  Springer, 2017, pp. 619--627.

\bibitem{merouche2015robotic}
S.~Merouche, L.~Allard, E.~Montagnon, G.~Soulez, P.~Bigras, and G.~Cloutier,
  ``A robotic ultrasound scanner for automatic vessel tracking and
  three-dimensional reconstruction of b-mode images,'' \emph{IEEE transactions
  on ultrasonics, ferroelectrics, and frequency control}, vol.~63, no.~1, pp.
  35--46, 2015.

\bibitem{Inrianadeau2016moments}
C.~Nadeau, A.~Krupa, J.~Petr, and C.~Barillot, ``Moments-based ultrasound
  visual servoing: From a mono-to multiplane approach,'' \emph{IEEE
  Transactions on Robotics}, vol.~32, no.~6, pp. 1558--1564, 2016.

\bibitem{pratt2015autonomousultrasound-guidedtissuedissection}
P.~Pratt, A.~Hughes-Hallett, L.~Zhang, N.~Patel, E.~Mayer, A.~Darzi, and G.-Z.
  Yang, ``Autonomous ultrasound-guided tissue dissection,'' in
  \emph{International Conference on Medical Image Computing and
  Computer-Assisted Intervention}.\hskip 1em plus 0.5em minus 0.4em\relax
  Springer, 2015, pp. 249--257.

\bibitem{Inriachevrie2019real}
J.~{Chevrie}, A.~{Krupa}, and M.~{Babel}, ``Real-time teleoperation of flexible
  beveled-tip needle insertion using haptic force feedback and 3d ultrasound
  guidance,'' in \emph{2019 International Conference on Robotics and Automation
  (ICRA)}, May 2019, pp. 2700--2706.

\bibitem{huang2018robotic}
Q.~Huang, J.~Lan, and X.~Li, ``Robotic arm based automatic ultrasound scanning
  for three-dimensional imaging,'' \emph{IEEE Transactions on Industrial
  Informatics}, vol.~15, no.~2, pp. 1173--1182, 2018.

\bibitem{zhang2017autonomous}
L.~Zhang, M.~Ye, P.~Giataganas, M.~Hughes, and G.-Z. Yang, ``Autonomous
  scanning for endomicroscopic mosaicing and 3d fusion,'' in \emph{2017 IEEE
  International Conference on Robotics and Automation (ICRA)}.\hskip 1em plus
  0.5em minus 0.4em\relax IEEE, 2017, pp. 3587--3593.

\bibitem{zhang2017macro}
L.~Zhang, M.~Ye, P.~Giataganas, M.~Hughes, A.~Bradu, A.~Podoleanu, and G.-Z.
  Yang, ``From macro to micro: Autonomous multiscale image fusion for robotic
  surgery,'' \emph{IEEE Robotics \& Automation Magazine}, vol.~24, no.~2, pp.
  63--72, 2017.

\bibitem{varghese2017framework}
R.~J. Varghese, P.~Berthet-Rayne, P.~Giataganas, V.~Vitiello, and G.-Z. Yang,
  ``A framework for sensorless and autonomous probe-tissue contact management
  in robotic endomicroscopic scanning,'' in \emph{2017 IEEE International
  Conference on Robotics and Automation (ICRA)}.\hskip 1em plus 0.5em minus
  0.4em\relax IEEE, 2017, pp. 1738--1745.

\bibitem{triantafyllou2018framework}
P.~Triantafyllou, P.~Wisanuvej, S.~Giannarou, J.~Liu, and G.-Z. Yang, ``A
  framework for sensorless tissue motion tracking in robotic endomicroscopy
  scanning,'' in \emph{2018 IEEE International Conference on Robotics and
  Automation (ICRA)}.\hskip 1em plus 0.5em minus 0.4em\relax IEEE, 2018, pp.
  2694--2699.

\bibitem{rosa2012building}
B.~Rosa, M.~S. Erden, T.~Vercauteren, B.~Herman, J.~Szewczyk, and G.~Morel,
  ``Building large mosaics of confocal edomicroscopic images using visual
  servoing,'' \emph{IEEE transactions on biomedical engineering}, vol.~60,
  no.~4, pp. 1041--1049, 2012.

\bibitem{giataganas2018intraoperative}
P.~Giataganas, M.~Hughes, C.~J. Payne, P.~Wisanuvej, B.~Temelkuran, and G.-Z.
  Yang, ``Intraoperative robotic-assisted large-area high-speed microscopic
  imaging and intervention,'' \emph{IEEE Transactions on Biomedical
  Engineering}, vol.~66, no.~1, pp. 208--216, 2018.

\bibitem{nikolaev2019ultrasound}
A.~Nikolaev, H.~H. Hansen, L.~de~Jong, R.~Mann, E.~Tagliabue, B.~Maris,
  V.~Groenhuis, F.~Siepel, M.~Caballo, I.~Sechopoulos, \emph{et~al.},
  ``Ultrasound-guided breast biopsy of ultrasound occult lesions using
  multimodality image co-registration and tissue displacement tracking,'' in
  \emph{Medical Imaging 2019: Ultrasonic Imaging and Tomography}, vol.
  10955.\hskip 1em plus 0.5em minus 0.4em\relax International Society for
  Optics and Photonics, 2019, p. 109550W.

\bibitem{op2011predicting}
J.~op~den Buijs, H.~H. Hansen, R.~G. Lopata, C.~L. de~Korte, and S.~Misra,
  ``Predicting target displacements using ultrasound elastography and finite
  element modeling,'' \emph{IEEE Transactions on Biomedical Engineering},
  vol.~58, no.~11, pp. 3143--3155, 2011.

\bibitem{daly2010fusion}
M.~J. Daly, H.~Chan, E.~Prisman, A.~Vescan, S.~Nithiananthan, J.~Qiu,
  R.~Weersink, J.~C. Irish, and J.~H. Siewerdsen, ``Fusion of intraoperative
  cone-beam ct and endoscopic video for image-guided procedures,'' in
  \emph{Medical Imaging 2010: Visualization, Image-Guided Procedures, and
  Modeling}, vol. 7625.\hskip 1em plus 0.5em minus 0.4em\relax International
  Society for Optics and Photonics, 2010, p. 762503.

\bibitem{de2010augmented}
L.~T. De~Paolis and G.~Aloisio, ``Augmented reality in minimally invasive
  surgery,'' in \emph{Advances in Biomedical Sensing, Measurements,
  Instrumentation and Systems}.\hskip 1em plus 0.5em minus 0.4em\relax
  Springer, 2010, pp. 305--320.

\bibitem{teber2009augmented}
D.~Teber, S.~Guven, T.~Simpfend{\"o}rfer, M.~Baumhauer, E.~O. G{\"u}ven,
  F.~Yencilek, A.~S. G{\"o}zen, and J.~Rassweiler, ``Augmented reality: a new
  tool to improve surgical accuracy during laparoscopic partial nephrectomy?
  preliminary in vitro and in vivo results,'' \emph{European urology}, vol.~56,
  no.~2, pp. 332--338, 2009.

\bibitem{grasa2011ekf}
O.~G. Grasa, J.~Civera, and J.~Montiel, ``Ekf monocular slam with
  relocalization for laparoscopic sequences,'' in \emph{2011 IEEE International
  Conference on Robotics and Automation}.\hskip 1em plus 0.5em minus
  0.4em\relax IEEE, 2011, pp. 4816--4821.

\bibitem{wang2015robot}
J.~Wang, L.~Qi, and M.~Q.-H. Meng, ``Robot-assisted occlusion avoidance for
  surgical instrument optical tracking system,'' in \emph{2015 IEEE
  International Conference on Information and Automation}.\hskip 1em plus 0.5em
  minus 0.4em\relax IEEE, 2015, pp. 375--380.

\bibitem{giannarou2016vision}
S.~Giannarou, M.~Ye, G.~Gras, K.~Leibrandt, H.~J. Marcus, and G.-Z. Yang,
  ``Vision-based deformation recovery for intraoperative force estimation of
  tool--tissue interaction for neurosurgery,'' \emph{International journal of
  computer assisted radiology and surgery}, vol.~11, no.~6, pp. 929--936, 2016.

\bibitem{giannarou2012probabilistic}
S.~Giannarou, M.~Visentini-Scarzanella, and G.-Z. Yang, ``Probabilistic
  tracking of affine-invariant anisotropic regions,'' \emph{IEEE transactions
  on pattern analysis and machine intelligence}, vol.~35, no.~1, pp. 130--143,
  2012.

\bibitem{mountney2007probabilistic}
P.~Mountney, B.~Lo, S.~Thiemjarus, D.~Stoyanov, and G.~Zhong-Yang, ``A
  probabilistic framework for tracking deformable soft tissue in minimally
  invasive surgery,'' in \emph{International Conference on Medical Image
  Computing and Computer-Assisted Intervention}.\hskip 1em plus 0.5em minus
  0.4em\relax Springer, 2007, pp. 34--41.

\bibitem{kazanzides-chen-etal-icra-2014dvrk}
P.~Kazanzides, Z.~Chen, A.~Deguet, G.~S. Fischer, R.~H. Taylor, and S.~P.
  DiMaio, ``An open-source research kit for the da vinci surgical system,'' in
  \emph{IEEE Intl. Conf. on Robotics and Auto. (ICRA)}, Hong Kong, China, 2014,
  pp. 6434--6439.

\bibitem{chaumette2006visualTut1}
F.~Chaumette and S.~Hutchinson, ``Visual servo control. i. basic approaches,''
  \emph{IEEE Robotics \& Automation Magazine}, vol.~13, no.~4, pp. 82--90,
  2006.

\bibitem{pratt2015robustUS}
P.~Pratt, A.~Jaeger, A.~Hughes-Hallett, E.~Mayer, J.~Vale, A.~Darzi, T.~Peters,
  and G.-Z. Yang, ``Robust ultrasound probe tracking: initial clinical
  experiences during robot-assisted partial nephrectomy,'' \emph{International
  journal of computer assisted radiology and surgery}, vol.~10, no.~12, pp.
  1905--1913, 2015.

\bibitem{shah2013solving}
M.~Shah, ``Solving the robot-world/hand-eye calibration problem using the
  kronecker product,'' \emph{Journal of Mechanisms and Robotics}, vol.~5,
  no.~3, p. 031007, 2013.

\bibitem{geiger2010efficient}
A.~Geiger, M.~Roser, and R.~Urtasun, ``Efficient large-scale stereo matching,''
  in \emph{Asian conference on computer vision}.\hskip 1em plus 0.5em minus
  0.4em\relax Springer, 2010, pp. 25--38.

\bibitem{rublee2011orb}
E.~Rublee, V.~Rabaud, K.~Konolige, and G.~R. Bradski, ``Orb: An efficient
  alternative to sift or surf.'' in \emph{ICCV}, vol.~11.\hskip 1em plus 0.5em
  minus 0.4em\relax Citeseer, 2011, p.~2.

\bibitem{kalal2010forward}
Z.~Kalal, K.~Mikolajczyk, and J.~Matas, ``Forward-backward error: Automatic
  detection of tracking failures,'' in \emph{2010 20th International Conference
  on Pattern Recognition}.\hskip 1em plus 0.5em minus 0.4em\relax IEEE, 2010,
  pp. 2756--2759.

\bibitem{garrido2014automatic}
S.~Garrido-Jurado, R.~Mu{\~n}oz-Salinas, F.~J. Madrid-Cuevas, and M.~J.
  Mar{\'\i}n-Jim{\'e}nez, ``Automatic generation and detection of highly
  reliable fiducial markers under occlusion,'' \emph{Pattern Recognition},
  vol.~47, no.~6, pp. 2280--2292, 2014.

\end{thebibliography}

\end{document}